\documentclass[10pt,twocolumn,letterpaper]{article}

\usepackage{cvpr}              

\usepackage{graphicx}
\usepackage{amsmath}
\usepackage{amssymb}
\usepackage{booktabs}
\usepackage[accsupp]{axessibility} 

\usepackage[pagebackref,breaklinks,colorlinks]{hyperref}

\usepackage[capitalize]{cleveref}
\crefname{section}{Sec.}{Secs.}
\Crefname{section}{Section}{Sections}
\Crefname{table}{Table}{Tables}
\crefname{table}{Tab.}{Tabs.}

\def\cvprPaperID{2} 
\def\confName{CVPR}
\def\confYear{2023}

\begin{document}

\title{Discriminative Spatial-Semantic VOS Solution: 1st Place Solution for 6th LSVOS}

\author{
    Deshui Miao\textsuperscript{1, 2}\thanks{Equal contribution.}, 
    Yameng Gu\textsuperscript{1}\footnotemark[1],  
    Xin Li\textsuperscript{2}\thanks{Team leader.}, 
    Zhenyu He\textsuperscript{1, 2}\thanks{Corresponding author.}, 
    Yaowei Wang\textsuperscript{2}, 
    Ming-Hsuan Yang\textsuperscript{3}\\
    \textsuperscript{1}Harbin Institute of Technology, Shenzhen \quad 
    \textsuperscript{2}Peng Cheng Laboratory\\ 
    \textsuperscript{3}University of California at Merced\\ 
    Team: HITSZ \& PCL VisionLab
}

\maketitle



\begin{abstract}
Video object segmentation (VOS) is a crucial task in computer vision, but current VOS methods struggle with complex scenes and prolonged object motions.
To address these challenges, the MOSE dataset aims to enhance object recognition and differentiation in complex environments, while the LVOS dataset focuses on segmenting objects exhibiting long-term, intricate movements. 
This report introduces a discriminative spatial-temporal VOS model that utilizes discriminative object features as query representations. 
The semantic understanding of spatial-semantic modules enables it to recognize object parts, while salient features highlight more distinctive object characteristics. 
Our model, trained on extensive VOS datasets, achieved first place (\textbf{80.90\%} $\mathcal{J \& F}$) on the test set of the 6th LSVOS challenge in the VOS Track, demonstrating its effectiveness in tackling the aforementioned challenges. 
The code will be available at \href{https://github.com/yahooo-m/VOS-Solution}{code}.
\end{abstract}


\begin{figure*}[t]
\centering
\includegraphics[width=1\textwidth]{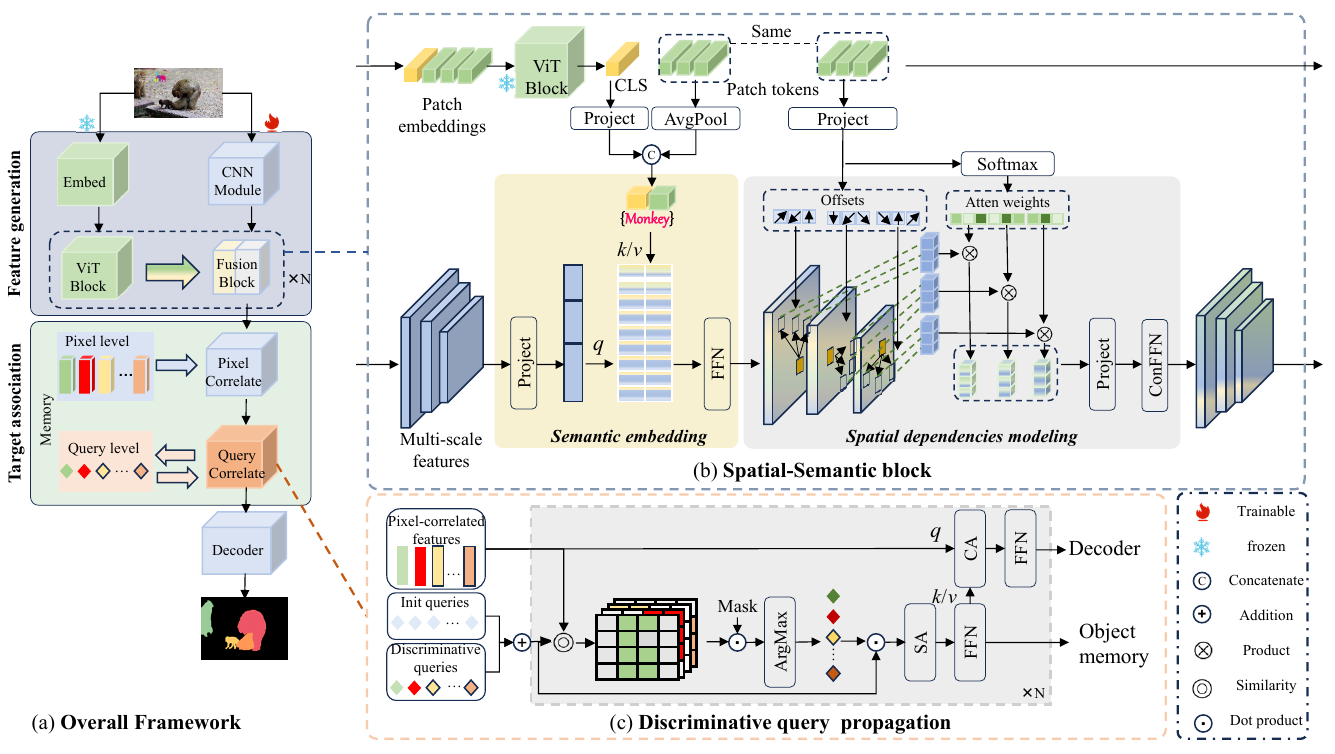}
\caption{Overall framework of our methods.}
\label{fig:overall}
\end{figure*} 
\section{Introduction}
Video Object Segmentation (VOS) aims to track and segment target objects in a video sequence, starting from mask annotations in the initial frame~\cite{ytvos2018, davis2017, feelvos, aot}. 
With the increasing volume of video data, VOS technology has broad applications in fields like autonomous driving, augmented reality, and interactive video editing~\cite{perclip, bao2018cnnmrf}. 
However, VOS tasks face numerous challenges, including drastic changes in object appearance, occlusions, and identity confusion caused by similar objects and background elements. 
These issues are particularly pronounced in long-term videos, further complicating the task.

VOS methods~\cite{STM, stcn} typically perform video object segmentation by comparing the test frame with previous frames. 
An association model is initially used to generate pixel-wise correlated features between the test frame and target templates. 
These correlated features are then utilized to accurately predict the target masks~\cite{SimVOS}. 
To handle the issue of target appearance changes over time, some methods~\cite{aot, xmem} incorporate a memory module that stores samples of the targets to capture their evolving appearances. 
Additionally, recent approaches~\cite{ISVOS, cutie} introduce object queries to differentiate various target objects, thereby reducing identity confusion.

Two datasets are proposed to evaluate the performance of current methods in realistic and complex environments involving complex or long-term motions. 
On one hand, the MOSE~\cite{mose} (coMplex video Object SEgmentation) dataset is a comprehensive collection designed to provide challenging scenarios with occluded and crowded objects. 
Unlike existing datasets that primarily feature relatively isolated and salient objects, MOSE focuses on realistic environments, highlighting the limitations of current VOS methods and encouraging the development of more robust algorithms. 
LVOS~\cite{Hong_2023_ICCV} (Long-term Video Object Segmentation) is the first densely annotated long-term video object segmentation dataset, which aims to provide a benchmark for the development and evaluation of long-term VOS models. 
The long-term videos in LVOS encompass multiple challenges, particularly attributes unique to long-term videos such as frequent reappearances and long-term similar object confusion.

Current methods face multiple challenges when handling the MOSE and LVOS datasets. 
First, when targets exhibit multiple complex or independent parts due to occlusion, background clutter, and shape complexity, existing methods tend to produce incomplete prediction masks. 
This is mainly because these methods rely too much on pixel-level correlation, focusing on pixel similarity while overlooking semantic information. 
Although object query representations~\cite{ISVOS, cutie} improve ID association accuracy, they perform poorly in sequences with significant target appearance changes and small targets. 
The current query propagation strategy, which updates using the entire online predicted target sample, easily introduces noise and accumulates errors, leading to tracking failures such as missing targets and ID switches, particularly when tracking small objects. 
Additionally, most existing methods are primarily designed for short-term VOS and struggle to handle long-term scenarios where understanding complex motions becomes increasingly challenging. 
These methods are prone to long-term disappearance and error accumulation, which can result in poor performance and even out-of-memory crashes due to the continuously expanding memory bank required for long videos.

To address the aforementioned issues, we propose a VOS framework to learn both the semantic prior, spatial dependence and discriminative query representation.
Specifically, we design a block that efficiently learns both semantic and detailed information, which can extract rich semantic features from a pre-trained Vision Transformer (ViT) without the need to train all feature extraction parameters.
To better model the query representation of the target, we design a more discriminative filtering mechanism to generate discriminative queries.

The main designs in this project include: 
\begin{itemize}
\item We propose a spatial-semantic block to incorporate the semantic information from the ViT feature for Video Object Segmentation. 
We use the cls token from a pretrained ViT backbone to extract the semantic prior and global average polling of the image patches to extract the global information of the current frames. 
Furthermore, we design a local fusion to leverage the spatial information from the ViT features. The proposed framework shows robustness when facing multiple complex sequences.
\item We develop a discriminative feature filtering module to model the discriminative query representation which achieves great improvement when dealing with small objects.
\end{itemize}

\section{Our solution}
To address the issues in VOS, we propose a robust video object segmentation method that incorporates semantic awareness and query enhancement. 
In this approach, we introduce the spatial-semantic module, which leverages the semantic and detailed information from pretrained ViT models to handle complex target appearance variations and ID confusion between visually similar targets. %
Specifically, we fuse the cls token information from ViT with multi-scale features and perform local fusion between frame patches and multi-scale features for detailed integration. 
Additionally, to ensure a stable representation of target queries, we develop a discriminative query representation module within the query transformer to capture the local representations of targets. We describe key components as follows, and for more details please refer to \cite{li2024learningspatialsemanticfeaturesrobust}. 

\subsection{Spatial-Semantic Block}
Since the VOS task involves generic objects without class labels, learning semantic representations directly from the VOS dataset during training is challenging. 
So we propose a spatial-semantic module to address the semantic understanding for video object segmentation tasks.

Figure~\ref{fig}(b) illustrates the spatial-semantic network, which comprises a semantic embedding module and a spatial dependency modeling module. 
The semantic embedding module integrates semantic information from a trained ViT model into multi-scale features, while the spatial dependency modeling module learns spatial dependencies based on these integrated features.

\subsection{Discriminative Query Generation}
We find that directly updating target query memory with complete object patches generated from online predicted masks is ineffective, as these predicted masks often contain background noise, making the target features less distinctive and causing errors to accumulate over time. 
To address this issue, we update target queries during frame propagation using the most discriminative features of the target object.

Specifically, by comparing the target query with each channel activation in the correlated feature map of the target and selecting the most similar one, we can extract the discriminative features of the target object. 
Based on this newly generated discriminative feature, we dynamically calculate the relationship between the salient query and salient pixel features in an additive manner to update the target queries. 
This discriminative query generation scheme adaptively refines target queries with the most representative features, helping to tackle the challenges of dramatic appearance variations in long-term videos.
\begin{table}
\centering
\caption{Ranked Top 4 results list for the joint test set from MOSE and LVOS. We mark our results in {\color{blue}{blue}}.}
\setlength{\tabcolsep}{2mm}{
\renewcommand\arraystretch{1.0}
\begin{tabular}{c | c | c c c}
\toprule
Rank & Team & $\mathcal{J}\&\mathcal{F}$ & $\mathcal{J}$ & $\mathcal{F}$ \\
\midrule
1 & \textcolor{blue}{Ours} &\textcolor{blue}{80.90} &\textcolor{blue}{76.16} &\textcolor{blue}{85.63} \\
2 & yuanjie &80.84 &76.42 &85.26\\
3 & Sch89.89 &76.35 &71.94 &80.76 \\
4 & MVP-TIME &75.79 &71.25 &80.33 \\

\bottomrule
\end{tabular}
}
\label{tab:performance}
\end{table}


\begin{figure*}[t]
\centering
\includegraphics[width=1\textwidth]{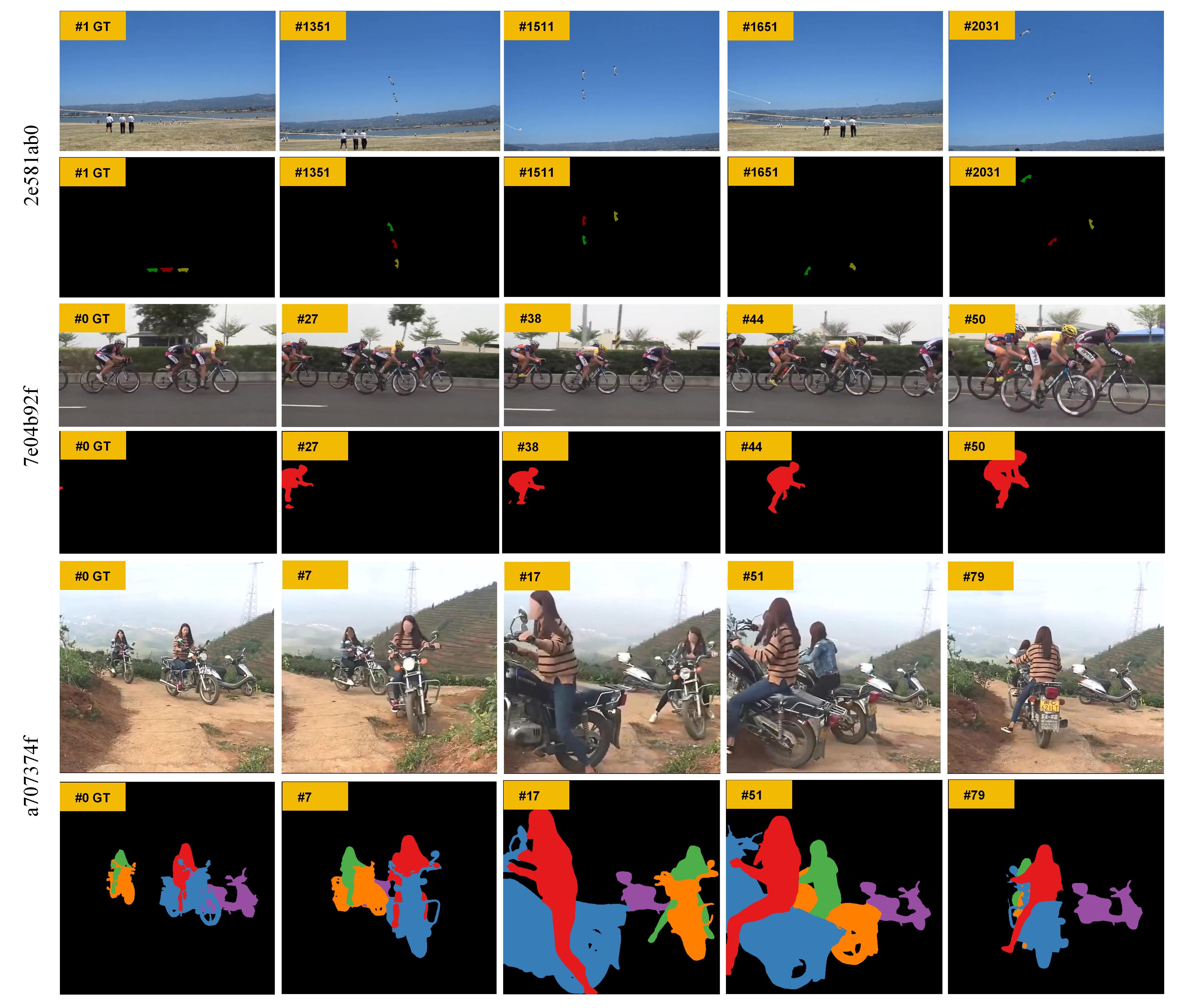} 
\vspace{-0.5cm}
\caption{Qualitative results on LSVOS sequences.}
\label{fig:result2}
\end{figure*} 

\section{Experiments}
\subsection{Challenge Description}
The 6th LSVOS challenge has two tracks: the VOS (Video Object Segmentation) track and the RVOS (Referring Video Object Segmentation) track. In this edition of the challenge, the classic YouTubeVOS~\cite{ytvos2018} benchmark used in the previous challenge is replaced by MOSE~\cite{mose} and LVOS~\cite{Hong_2023_ICCV} to study the VOS under more challenging complex environments. MOSE~\cite{mose} focuses on complex scenes, including the disappearance-reappearance of objects, inconspicuous small objects, heavy occlusions, crowded environments, etc. LVOS~\cite{Hong_2023_ICCV} focuses on long-term videos, with complex object motion and long-term reappearance. Besides, the origin YouTubeRVOS~\cite{URVOS} benchmark is replaced by MeViS~\cite{mevis}. MeViS~\cite{mevis} focuses on referring the target object in a video through its motion descriptions instead of static attributes, which breaks the basic design principles behind existing RVOS methods and boosts the rethinking of motion modeling.

\subsection{Implementation Details}

\noindent \textbf{Training}. 
Our training settings are similar to Cutie~\cite{cutie}. 
To enhance the performance of our model, we combine the MEGA dataset constructed by Cutie, which includes the YouTubeVOS~\cite{ytvos2018}, DAVIS~\cite{davis2017}, OVIS~\cite{ovis}, MOSE~\cite{mose}, and BURST~\cite{burst} datasets, with LaSOT~\cite{fan2020lasothighqualitylargescalesingle} dataset.
We sample eight frames to train the model, and three are randomly selected to train the matching process.
For each sequence, we randomly choose at most three targets for training.
The point supervision in loss is adopted to reduce the memory requirements.
We train the model for 195k on the joint MEGA and LaSOT dataset.
All our models are trained on 8 x NVIDIA V100 GPUs and tested on an NVIDIA A100 GPU.

\noindent\textbf{Inference.} 
Our inference is based on the released challenge dataset, which combines MOSE and LVOS test datasets.
Our feature and query memory is updated every 3rd frame during the testing phase. 
For longer sequences, we employ a long-term fusion strategy~\cite{xmem} for updating. To enhance storage quality, we skip frames without targets and do not store them.
The test input size contains two scales: 720 for general size and 1080 for small targets.
The final score is a version of multi-scale fusion~\cite{xmem}.

\noindent \textbf{Evaluation Metrics.} 
We use mean Jaccard $\mathcal{J}$ index and mean boundary
$\mathcal{F}$ score, along with mean $\mathcal{J}\&\mathcal{F}$ to evaluate segmentation accuracy. 

\subsection{Results}
The proposed method secures 1st place on the VOS track of the 6th LSVOS Challenge 2024, as shown in Table~\ref{tab}. Additionally, Figure \ref{fig} presents some of our quantitative results. It is evident that the proposed solution accurately segments small targets and distinguishes similar targets in challenging scenarios characterized by severe appearance changes, and confusion among multiple similar objects, and small targets. These results also validate our method’s ability to capture complex motions and reappearances over the long term. Across the five submission versions, we observe that certain inference parameters affect performance, including test size, memory interval, the presence or absence of memory, flip augmentation, and multi-scale fusion.









\section{Conclusion}
\label{sec:Conclusion}
In this paper, we propose a robust solution for the task of video object segmentation, which helps the model understand the semantic information of the targets and generate discriminative queries of the target. 
In the end, we achieve 1st place on the VOS track of the 6th LSVOS Challenge 2024 on complex scene and long-term complex motion object segmentation coalescent task with 80.90\% $\mathcal{J}\&\mathcal{F}$.
We have released the full version of our method, and the code will also be released soon.
{\small
\bibliographystyle{plain}
\bibliography{egbib}
}

\end{document}